\documentclass[journal]{IEEEtran}

\usepackage[T1]{fontenc}
\usepackage{newtxtext}
\usepackage{amsthm}
\usepackage{newtxmath}
\usepackage{microtype}

\usepackage{mathtools,bm}
\usepackage{algorithm}
\usepackage{algpseudocode}

\usepackage{graphicx}
\usepackage{booktabs}
\usepackage{siunitx}
\usepackage{enumitem}
\usepackage{flushend}
\usepackage{xcolor}
\usepackage{csquotes}

\usepackage{xurl}
\usepackage{hyperref}
\usepackage{bookmark}
\usepackage{zref-check}
\usepackage{zref-clever}
\zcsetup{
  cap=true,
  abbrev=false,
  nameinlink=true,
  sort=true,
  comp=true
}

\newtheorem{theorem}{Theorem}[section]
\newtheorem{lemma}{Lemma}[section]
\newtheorem{proposition}{Proposition}[section]
\newtheorem{corollary}{Corollary}[section]
\newtheorem{assumption}{Assumption}[section]
\theoremstyle{definition}
\newtheorem{definition}{Definition}[section]

\newtheorem{counterexample}{Counterexample}
\theoremstyle{remark}

\theoremstyle{plain}

\zcRefTypeSetup{theorem}{
  Name-sg=Theorem,
  name-sg=theorem,
  Name-pl=Theorems,
  name-pl=theorems
}
\zcRefTypeSetup{lemma}{
  Name-sg=Lemma,
  name-sg=lemma,
  Name-pl=Lemmas,
  name-pl=lemmas
}
\zcRefTypeSetup{proposition}{
  Name-sg=Proposition,
  name-sg=proposition,
  Name-pl=Propositions,
  name-pl=propositions
}
\zcRefTypeSetup{corollary}{
  Name-sg=Corollary,
  name-sg=corollary,
  Name-pl=Corollaries,
  name-pl=corollaries
}
\zcRefTypeSetup{assumption}{
  Name-sg=Assumption,
  name-sg=assumption,
  Name-pl=Assumptions,
  name-pl=assumptions
}
\zcRefTypeSetup{definition}{
  Name-sg=Definition,
  name-sg=definition,
  Name-pl=Definitions,
  name-pl=definitions
}
\zcRefTypeSetup{example}{
  Name-sg=Example,
  name-sg=example,
  Name-pl=Examples,
  name-pl=examples
}
\zcRefTypeSetup{counterexample}{
  Name-sg=Counterexample,
  name-sg=counterexample,
  Name-pl=Counterexamples,
  name-pl=counterexamples
}
\zcRefTypeSetup{remark}{
  Name-sg=Remark,
  name-sg=remark,
  Name-pl=Remarks,
  name-pl=remarks
}
\zcRefTypeSetup{algorithm}{
  Name-sg=Algorithm,
  name-sg=algorithm,
  Name-pl=Algorithms,
  name-pl=algorithms
}

\newcommand{\R}{\mathbb{R}}
\newcommand{\cG}{\mathcal{G}}
\newcommand{\cR}{\mathcal{R}}
\newcommand{\cH}{\mathcal{H}}
\newcommand{\cC}{\mathcal{C}}
\newcommand{\one}{\mathbf{1}}
\newcommand{\eps}{\varepsilon}
\newcommand{\sgn}{\operatorname{sgn}}
\newcommand{\TV}{\operatorname{TV}}
\newcommand{\Reg}{\operatorname{Reg}}

\newcommand{\inner}[2]{\left\langle#1,#2\right\rangle}
\newcommand{\norm}[1]{\left\lVert#1\right\rVert}
\newcommand{\abs}[1]{\left\lvert#1\right\rvert}

\newcommand{\KinkCorrectedTerminal}{-13.027}
\newcommand{\KinkIndependentTerminal}{7.064\times10^{79}}

\newcommand{\GossipCounterexampleHorizon}{300}
\newcommand{\GossipBaselineTerminal}{1.391\times10^{15}}
\newcommand{\GossipBaselineLocalTerminal}{-8.334\times10^{16}}

\newcommand{\ExperimentHorizon}{12,000}
\newcommand{\ExperimentSeeds}{5}

\newcommand{\RealWorldNodes}{10}
\newcommand{\RealWorldSeeds}{5}
\newcommand{\RealWorldEndowment}{1}

\title{Black-Box Comparator-Adaptive Network Regret}
\hypersetup{
  pdftitle={Black-Box Comparator-Adaptive Network Regret},
  pdfauthor={Tomas Ortega, Hamid Jafarkhani}
}

\author{
    Tomas~Ortega,~\IEEEmembership{Graduate~Student~Member,~IEEE,}
    and~Hamid~Jafarkhani,~\IEEEmembership{Fellow,~IEEE}%
    \thanks{This work was supported in part by the NSF Award ECCS-2207457.}
    \thanks{Tomas Ortega was with the Department of Electrical Engineering and Computer Science, University of California, Irvine, Irvine, CA 92697 USA\@. He is now with Princeton Language and Intelligence (PLI), Princeton University, Princeton, NJ 08544 USA (e-mail: tomas.ortega@princeton.edu).}%
    \thanks{Hamid Jafarkhani is with the Department of Electrical Engineering and Computer Science, University of California, Irvine, Irvine, CA 92697 USA (e-mail: hamidj@uci.edu).}%
}

\begin{document}
\maketitle

\begin{abstract}
  We study comparator-adaptive decentralized online learning with network regret, which evaluates each node's prediction on the loss of every node in the network.
  We show that existing gossip-based algorithms are fundamentally limited for this task, and cannot achieve sublinear network regret with only one gossip step per decision round.
  To prove this, we construct a two-node counterexample, which shows that gossip-based algorithms can produce exponential regret despite bounded disagreement, resolving a previously open question in the literature.
  We then introduce a black-box reduction, which uses local learners with sublinear comparator-adaptive regret to obtain sublinear comparator-adaptive network regret.
  Each node exchanges predictions with its neighbors once per decision round, and a universal choice of edge weights works on every connected undirected graph without graph preprocessing.
  We also characterize admissible edge weights and show how weights adapted to the graph improve the regret bound.
  For example, for the complete graph, this improvement is quadratic in the number of nodes.
  Experiments on synthetic and real-world data illustrate the counterexamples we propose and compare the reduction with decentralized baselines.
\end{abstract}

\section{Introduction}\label{sec:introduction}

Much of the literature on online learning studies a centralized setting~\cite{hazan2016introduction,orabona2019modern}.
A single learner repeatedly makes predictions and observes convex losses and their subgradients at the predictions.
The learner's performance is measured by regret, which compares the cumulative loss of the learner's predictions to that of a fixed comparator in hindsight.
However, in many applications, multiple nodes hold the data that define the problem, and each node observes only its own local loss.
This decentralized setting, with its own notion of regret, is less explored.

To formally define the problem, consider a network of $N$ nodes, where each node
repeatedly issues predictions and observes only its own local loss.
We denote the prediction of node $n$ at round $t$ by $x_{n,t}$, and the local loss function by $\ell_{n,t}$.
A first natural extension of the single-node regret notion is to evaluate each node's prediction on its own local loss.
This is the \emph{local regret}, and we formally define the one-round and cumulative local regrets by
\begin{align*}
  r_t^{\mathrm{loc}}(u)
   & \coloneqq
  \frac1N\sum_{n=1}^N
  \left(\ell_{n,t}(x_{n,t})-\ell_{n,t}(u)\right), \\
  R_T^{\mathrm{loc}}(u)
   & \coloneqq\sum_{t=1}^T r_t^{\mathrm{loc}}(u),
\end{align*}
where $T$ is a fixed number of rounds and $u$ is any comparator in the decision space.
However, this does not exploit the benefits of having multiple nodes, as each node could ignore the predictions of the others and still achieve sublinear local regret: each node can run a single-node online learning algorithm independently, which are well-studied and can achieve sublinear regret~\cite{hazan2016introduction,orabona2019modern}.
Therefore, we consider a setting in which a prediction produced at one node may later be evaluated on data from every other node.
This cross-evaluation indicates whether nodes are learning from the data from nodes in the entire network, not just their own.
Therefore, the natural common-comparator regret contains every cross-evaluation
$\ell_{m,t}(x_{n,t})$, instead of just $\ell_{n,t}(x_{n,t})$.
We call this quantity \emph{network regret}.
To the best of our knowledge, Koppel et~al.~\cite{koppel2015saddle} were the first to study this notion, but they did so on bounded domains and without comparator-adaptive guarantees.
Formally, for a fixed comparator $u$, we let the one-round and cumulative network
regrets be, respectively,
\begin{align*}
  r_t^{\mathrm{net}}(u)
   & \coloneqq
  \frac1{N^2}\sum_{n=1}^N\sum_{m=1}^N
  \left(\ell_{m,t}(x_{n,t})-\ell_{m,t}(u)\right), \\
  R_T^{\mathrm{net}}(u)
   & \coloneqq\sum_{t=1}^T r_t^{\mathrm{net}}(u).
\end{align*}

For most of the exposition, we use a real-valued decision space and real-valued local losses, as this will ease the presentation.
We extend our results to Hilbert spaces in \zcref{sec:extensions}.

Previous work on decentralized comparator-adaptive online learning proposed the use of \emph{gossip} algorithms, that is, algorithms that communicate and average the internal states of local learners~\cite{ortega2025deco,ortega2026decoef}.
Regret guarantees derived in this line of work need bounds on the state agreement between nodes to turn them into bounds on the network regret.
Notably, previous research showed that if the decentralized algorithm DECO-II uses a linearly growing number of gossip steps per decision round, then the network regret is sublinear.
Whether a constant number of gossip steps per decision round suffices to guarantee sublinear network regret was left as an open question~\cite{ortega2025deco}.
In this work, we answer this question in the negative for the case of a single gossip step per decision round.

To overcome this fundamental limitation of existing gossip algorithms, we propose a black-box reduction that changes the feedback of each local learner.
It turns sublinear comparator-adaptive local regret into sublinear comparator-adaptive network regret.

The main contributions of this work are as follows:
\begin{enumerate}%
  \item \textbf{Resolving an open question from gossip-based algorithms.}
        It was conjectured that the gossip-based algorithm DECO-II, with a constant number of gossip steps per decision round, can guarantee sublinear network regret~\cite{ortega2025deco}.
        We show that for a single round of gossip per decision round, the conjecture is not true. We construct a counterexample in which the network regret is exponential in the number of rounds.
        This shows the limitations of existing gossip algorithms and motivates our black-box reduction.
  \item \textbf{A preprocessing-free black-box reduction.}
        We consider a scenario where a network of nodes run online linear learning algorithms, and each of them has a sublinear comparator-adaptive network regret for bounded feedback.
        This property is satisfied by well-known existing algorithms~\cite{orabona2016coin,orabona2019modern,mcmahan2014unconstrained}.
        We then construct a black-box reduction that turns these local algorithms into a decentralized algorithm with sublinear comparator-adaptive network regret.
        We show that this can be done with one communication round per decision round and no graph preprocessing.

  \item \textbf{Topology-adaptive edge weights.}
        We show how to refine the communication graph's edge weights to improve the network regret bound constants.
        We characterize the admissible edge weights and discuss different weight designs.
\end{enumerate}

\paragraph*{Organization}
The remainder of the paper is organized as follows.
We review prior work in \zcref{sec:literature} and define the problem in \zcref{sec:model}.
In \zcref{sec:reduction}, we develop the reduction for any given edge weights, and later present a universal weight choice that does not need graph preprocessing.
\zcref{sec:kt-analysis} applies it to the Krichevsky-Trofimov (KT) learner, a well-known comparator-adaptive online learning algorithm.
The counterexamples in \zcref{sec:counterexamples} explain why local regret bounds and one gossip step cannot control network regret.
We then study weight choices in \zcref{sec:graph-factors} and show how to extend the reduction from single-dimensional to multi-dimensional predictions in a Hilbert space in \zcref{sec:extensions}.
\zcref{sec:experiments} presents experiments on synthetic and real-world data, among them a realization of the two-node counterexample, and \zcref{sec:conclusion} concludes the paper.
\section{Related Work}\label{sec:literature}

\subsection{Comparator-adaptive online learning}

Classic online learning algorithms require a known bound on the comparator norm to achieve sublinear regret.
Comparator-adaptive online learning, also sometimes called parameter-free online learning, removes this requirement~\cite{streeter2012no,mcmahan2014unconstrained,orabona2015scale}.
We can analyze comparator-adaptive online learning by analyzing an associated coin betting game: Orabona and Pál~\cite{orabona2016coin} showed that we can convert wealth guarantees in a coin betting game into comparator regret through the convex conjugate.
They also introduced the KT algorithm, named after the Krichevsky-Trofimov estimator, which is a simple comparator-adaptive online learning algorithm that achieves sublinear regret, and which we will use as a base learner for our experimental results in~\zcref{sec:experiments}.
Further work extended these algorithms to Banach spaces, developed second-order guarantees, and made them adaptive to unknown subgradient scales~\cite{cutkosky2018blackbox,mhammedi2020lipschitz,cutkosky2020dynamic,jacobsen2022parameterfree}.
Among these works, our black-box reduction is closest to Cutkosky and Orabona~\cite{cutkosky2018blackbox}, as our proof also relies on transformations of the base learners' inputs, but in our case to make them amenable to a network setting.

\subsection{Decentralized online learning}
Distributed subgradient and dual-averaging methods use network averaging for offline and online optimization~\cite{nedic2009distributed,duchi2012dual,yan2013distributed}.
Gossip algorithms exchange and average information between neighboring nodes~\cite{xiao2004fastest,boyd2006random}.
Koppel et~al.~\cite{koppel2015saddle} control network regret on bounded domains with a distributed saddle-point method.
Later work treats changing directed graphs, dynamic environments, constraints, and network-dependent regret~\cite{mateos2014distributed,shahrampour2018distributed,yi2020distributed}.
Communication compression is also considered in the literature~\cite{tu2022compressed,zhou2026revisiting,shen2021distributed,ortega2023gossiped}.
To improve the dependence on the network size and connectivity, accelerated decentralized follow-the-regularized-leader schemes have been proposed~\cite{wan2024nearly,wan2025optimal}.

Prior work has also studied several notions of regret: a good survey can be found in Hsieh et~al.~\cite{hsieh2022multi}.
For one of the notions of regret, called joint regret, the authors consider a setting in which only one node is active per round.
For this setting, Hoeven et~al.~\cite{vanderhoeven2022joint} give a comparator-adaptive algorithm with sublinear joint regret where the nodes have a bounded communication budget.

Achddou et~al.~\cite{achddou2024availability} study network regret under stochastic agent availability and propose Gossip-FTRL, which is not comparator-adaptive.
More recent gossip methods obtain comparator-adaptive network regret guarantees, including with compressed communication~\cite{ortega2025deco,ortega2026decoef}.
Their guarantees require a linearly growing number of gossip steps per decision round.

\section{Problem setup}\label{sec:model}

Let $\cG=(V,E)$ be a fixed connected undirected graph with $V\coloneqq\{1,\ldots,N\}$ and $N\ge1$.
Two nodes are connected if and only if they can communicate.
Communication occurs in rounds $t=1,2,\ldots,T$, and is synchronous for all agents at each round.

In Round $t$, Node $n$ predicts $x_{n,t}\in\R$.
We stack the predictions in the vector $x_t\coloneqq{(x_{n,t})}_{n\in V}$, which has dimension $N$.
After the nodes reveal the prediction vector $x_t$, an adversary selects finite convex losses $\ell_{n,t}:\R\to\R$.
Node $n$ observes a subgradient $g_{n,t}$ of its local loss at $x_{n,t}$, meaning that
\begin{equation}
  \ell_{n,t}(x_{n,t})+g_{n,t}(y-x_{n,t})\leq\ell_{n,t}(y)
  \qquad\text{for all }y\in\R.
  \label{eq:local-subgradient}
\end{equation}
We use $\partial\ell_{n,t}(x)$ for the set of all subgradients of $\ell_{n,t}$ at $x$.

\begin{assumption}[Uniform subgradient bound]\label{ass:gradient-bound}
  A known constant $G>0$ satisfies
  \begin{equation*}
    \sup_{x\in\R}\sup_{g\in\partial\ell_{n,t}(x)}\abs g
    \le G
    \qquad\text{for every }n,t.
  \end{equation*}
\end{assumption}
We need a global bound on the subgradients to control a loss when another node's prediction is substituted into it.
However, a distributed algorithm is desired to use only the local subgradients in \zcref{eq:local-subgradient}, as done in this work.
By Shalev-Shwartz~\cite[Lemma~2.6]{shalev2012online}, \zcref{ass:gradient-bound} results in
\begin{equation}
  \abs{\ell_{n,t}(x)-\ell_{n,t}(y)}
  \le G\abs{x-y}
  \qquad\text{for all }x,y\in\R.
  \label{eq:gradient-difference-bound}
\end{equation}

Let us define the \emph{complete all-pairs disagreement} of the prediction vector $x$ as
\begin{equation}
  D_N(x)
  \coloneqq
  \frac1{N^2}\sum_{n=1}^N\sum_{m=1}^N\abs{x_n-x_m}.\label{eq:disagreement}
\end{equation}
This is used to upper bound the network regret in terms of local regret and prediction disagreement as follows: adding and subtracting $\ell_{m,t}(x_{m,t})$ in each summand of $r_t^{\mathrm{net}}(u)$, then using \zcref{eq:gradient-difference-bound}, gives
\begin{equation}
  r_t^{\mathrm{net}}(u)
  \le r_t^{\mathrm{loc}}(u)+G D_N(x_t).
  \label{eq:network-local-bound}
\end{equation}
The first term corresponds to local regret and the second term is the network disagreement term.
Applying convexity to each local loss at $x_{n,t}$ gives the basic linearization
\begin{equation}
  r_t^{\mathrm{net}}(u)
  \le
  \frac1N\sum_{n=1}^N g_{n,t}(x_{n,t}-u)
  +G D_N(x_t).
  \label{eq:network-linearization}
\end{equation}
This linearization was also used in prior work to analyze gossip-based algorithms~\cite{ortega2025deco,ortega2026decoef}, and previously by Hsieh et~al.~\cite{hsieh2022multi} to analyze different notions of regret.

\begin{definition}[Bounded-feedback online linear learner]\label{def:base-learner}
  A learner $\mathsf A_n$ outputs a decision $a_{n,t}\in\R$ and then receives a scalar feedback value $z_{n,t}\in[-1,1]$.
  It has a regret bound $\Reg_{n,T}(v)$ that satisfies
  \begin{equation*}
    \sum_{t=1}^T z_{n,t}(a_{n,t}-v)
    \le \Reg_{n,T}(v)
    \qquad\text{for every }v\in\R.
  \end{equation*}
\end{definition}
A desirable property of this definition is that it allows us to analyze the decentralized algorithm independently of the base learners, using only their regret guarantees $\Reg_{n,T}$.
The algorithm interacts with each learner $\mathsf A_n$ only through \Call{Predict}{$\mathsf A_n$} and \Call{Update}{$\mathsf A_n,z_{n,t}$}, unlike gossip-based approaches, which need to access and communicate the learners' internal states.

\section{The black-box reduction}\label{sec:reduction}

The network-regret bound in \zcref{eq:network-linearization} contains prediction disagreement, and prior work used gossip to obtain upper bounds on this term.
In this work, we instead use a graph total variation (or graph-TV) penalty to dominate this term.
Related penalties are used for estimation along sequences~\cite{wahlberg2012admm}, signal and image processing on multidimensional grids~\cite{yang2013efficient}, and clustering and optimization on general graphs in the network lasso~\cite{hallac2015network}.
In the optimization literature, Cao and Başar~\cite{cao2021signs} use sign-based neighbor corrections and total-variation regularized online losses, but their analysis does not provide comparator-adaptive guarantees.
In our reduction, we add a correction to the local feedback at each node for each incident edge, which allows us to use the graph-TV bound.
In this section, we present the main results of the paper: we first characterize the required weights on the graph edges, then give the black-box reduction, and finally a universal weight choice that needs no graph preprocessing.

\subsection{Weighted total variation and cut domination}

Let $\beta={(\beta_e)}_{e\in E}$ be nonnegative edge weights.
If some edges in the graph have zero weight, they do not require online communication.
The weighted graph total variation is
\begin{equation*}
  \TV_\beta(x)
  \coloneqq
  \sum_{\{i,j\}\in E}\beta_{\{i,j\}}\abs{x_i-x_j}.
\end{equation*}
We need these weights to make the graph-TV bound the all-pairs disagreement from \zcref{eq:disagreement}.
Next, we present a cut-based characterization of the admissible weights.
Similar ideas appear in the literature on graph total variation inequalities~\cite{tillich2000edge} and max-flow/min-cut theory~\cite{leighton1999multicommodity}.

For $U\subseteq V$, the cut $\delta(U)$ is the set of edges that connect nodes in $U$ to nodes in $V\setminus U$.
Its total weight is
\begin{equation*}
  \beta(\delta(U))
  \coloneqq
  \sum_{e\in\delta(U)}\beta_e.
\end{equation*}

\begin{definition}[Admissible edge weights]\label{def:cut-certificate}
  We call the weights $\beta$ admissible if all nonempty proper subsets $U\subset V$ satisfy
  \begin{equation}
    \beta(\delta(U))
    \ge
    \frac{2\abs U(N-\abs U)}{N^2}.
    \label{eq:cut-domination}
  \end{equation}
\end{definition}

Observe that there are $\abs U(N-\abs U)$ pairs of nodes with one node in each group.
In the complete graph, each pair is connected by an edge.
And, if each edge of the complete graph has weight $2/N^2$, the total weight of this cut is exactly the right-hand side of \zcref{eq:cut-domination}.
Therefore, admissibility requires that every cut in $\cG$ have at least the weight of the same cut in the complete graph with edges of weight $2/N^2$.

Now, we can characterize the admissible weights in terms of the total variation of the graph.
This can be derived, with some work, from the equivalence results between graph total variation inequalities and cut inequalities~\cite[Theorem~2 and Example~2]{tillich2000edge}, but we include a self-contained proof for completeness.

\begin{lemma}[Exact cut characterization]\label{lem:cut-characterization}
  The weights $\beta$ are admissible if and only if
  \begin{equation}
    D_N(x)\le\TV_\beta(x)
    \qquad\text{for every }x\in\R^N.
    \label{eq:pairwise-tv-domination}
  \end{equation}
\end{lemma}
\begin{IEEEproof}
  For necessity, fix a nonempty proper subset $U\subset V$ and set $x=\one_U$, which assigns value $1$ to nodes in $U$ and $0$ to all other nodes.
  Note that only node pairs on opposite sides of the cut contribute to $D_N(x)$, and only cut edges contribute to $\TV_\beta(x)$, which implies
  \begin{align*}
    D_N(\one_U)       & =\frac{2\abs U(N-\abs U)}{N^2}, \\
    \TV_\beta(\one_U) & =\beta(\delta(U)).
  \end{align*}
  Substituting into \zcref{eq:pairwise-tv-domination} gives \zcref{eq:cut-domination}.

  For sufficiency, assume the weights are admissible and define $U_\tau=\{n:x_n>\tau\}$.
  Two nodes, with values $x_i$ and $x_j$, lie on the opposite sides of $U_\tau$ exactly when $\tau$ lies between their values.
  All possible thresholds $\tau$ that separate the two nodes form an interval of length $\abs{x_i-x_j}$.
  Note that we can express each interval length as an integral over the indicator function of the interval, which is $1$ when $\tau$ lies between $x_i$ and $x_j$, and $0$ otherwise.

  For each threshold, there are $2\abs{U_\tau}(N-\abs{U_\tau})$ ordered pairs of nodes on opposite sides, and the edges crossing between the two sides have total weight $\beta(\delta(U_\tau))$.
  Summing over pairs and edges, and using the indicator function trick, gives
  \begin{align*}
    D_N(x)
     & =\frac1{N^2}\sum_{n=1}^N\sum_{m=1}^N\abs{x_n-x_m}      \\
     & =\int_{-\infty}^{\infty}
    \frac{2\abs{U_\tau}(N-\abs{U_\tau})}{N^2}\,\mathrm d\tau, \\
    \TV_\beta(x)
     & =\sum_{\{i,j\}\in E}\beta_{\{i,j\}}\abs{x_i-x_j}       \\
     & =\int_{-\infty}^{\infty}
    \beta(\delta(U_\tau))\,\mathrm d\tau.
  \end{align*}
  Note that both quantities are zero when $U_\tau$ is empty or equal to $V$.
  Otherwise, we can upper bound the first integrand by the second thanks to the admissibility of the weights, and integrating proves \zcref{eq:pairwise-tv-domination}.
\end{IEEEproof}

\subsection{The local algorithm}

For each node, define the \emph{local incident load} and \emph{normalization scale} by
\begin{align*}
  \kappa_n
   & \coloneqq N\sum_{e\ni n}\beta_e,
   &
  \lambda_n
   & \coloneqq G(1+\kappa_n).
\end{align*}
In other words, $\kappa_n$ is the sum of the weights of the edges incident to Node $n$, scaled by $N$, and $\lambda_n$ is the local scale used to normalize the prediction $a_{n,t}$.
For convenience, we write $\bm\kappa={(\kappa_n)}_{n\in V}$ and $\bm\lambda={(\lambda_n)}_{n\in V}$.

Let $a_t,x_t,g_t,c_t,z_t\in\R^N$ denote the quantities of the stacked node.
The correction at Node $n$ is
\begin{equation*}
  c_{n,t}
  \coloneqq GN\sum_{j:\{n,j\}\in E}
  \beta_{\{n,j\}}\sgn(x_{n,t}-x_{j,t}),
\end{equation*}
where $\sgn(0)\coloneqq0$.
The predictions and normalized feedback are
\begin{align*}
  x_{n,t}
   & \coloneqq\frac{a_{n,t}}{\lambda_n},            \\
  \widetilde g_t
   & \coloneqq g_t+c_t,
   &
  z_{n,t}
   & \coloneqq\frac{\widetilde g_{n,t}}{\lambda_n}.
\end{align*}

\begin{algorithm}[t]
  \caption{Edge-corrected learning at node $n$}\label{alg:local}
  \begin{algorithmic}[1]
    \Require{} Learner $\mathsf A_n$, bound $G$, network size $N$,
    incident edges and their weights, local scale $\lambda_n$
    \State{} Initialize the private learner $\mathsf A_n$
    \For{$t=1,2,\ldots$}
    \State{} $a_{n,t}\gets\Call{Predict}{\mathsf A_n}$
    \State{} $x_{n,t}\gets a_{n,t}/\lambda_n$ and predict $x_{n,t}$
    \State{} Observe any $g_{n,t}\in\partial\ell_{n,t}(x_{n,t})$
    \State{} Exchange $x_{n,t}$ with neighbors $j$ with $\beta_{\{n,j\}}>0$
    \State{} $c_{n,t}\gets0$
    \ForAll{incident edges $e=\{n,j\}$ with $\beta_e>0$}
    \State{} $c_{n,t}\gets c_{n,t}
      +GN\beta_e\sgn(x_{n,t}-x_{j,t})$
    \EndFor{}
    \State{} $z_{n,t}\gets(g_{n,t}+c_{n,t})/\lambda_n$
    \State{} $\Call{Update}{\mathsf A_n,z_{n,t}}$
    \EndFor{}
  \end{algorithmic}
\end{algorithm}

\begin{lemma}[Correction identities and local boundedness]\label{lem:correction-identities}
  For every comparator $u$,
  \begin{align*}
    \one^\top c_t
     & =0,               \\
    \frac1N c_t^\top(x_t-u\one)
     & =G\TV_\beta(x_t), \\
    \abs{c_t}
     & \le G\bm\kappa.
  \end{align*}
  The final inequality is componentwise, so $\abs{z_t}\le\one$ componentwise as well.
\end{lemma}
\begin{IEEEproof}
  For the first identity, note that each edge contributes opposite corrections at its two endpoints, so the sum of all corrections is zero.
  For the second identity, we expand $N^{-1}c_t^\top(x_t-u\one)$ as the sum over all edges of $G\beta_{\{i,j\}}\abs{x_{i,t}-x_{j,t}}$, where we have used the first identity to cancel the $u$ term, and the results follow.
  For the third expression, the triangle inequality gives
  \begin{equation*}
    \abs{c_{n,t}}\le GN\sum_{e\ni n}\beta_e=G\kappa_n.
  \end{equation*}
  Since $\abs{g_{n,t}}\le G$, we also have
  \begin{equation*}
    \abs{z_{n,t}}\le\frac{G(1+\kappa_n)}{\lambda_n}=1.
    \IEEEQEDhereeqn
  \end{equation*}
\end{IEEEproof}

\begin{theorem}[Black-box network regret bound]\label{thm:main}
  Given admissible edge weights $\beta$ and under \zcref{ass:gradient-bound}, for every time horizon, comparator, loss sequence, and valid local subgradients, \zcref{alg:local} with base learners satisfying \zcref{def:base-learner} satisfies
  \begin{equation*}
    R_T^{\mathrm{net}}(u)
    \le
    \frac1N\sum_{n=1}^N\Reg_{n,T}(\lambda_n u).
  \end{equation*}
\end{theorem}
\begin{IEEEproof}
  We start from the upper bound of network regret in terms of the local regret and the disagreement term in \zcref{eq:network-local-bound}.
  Since $\beta$ is admissible, by \zcref{lem:cut-characterization}, we obtain
  \begin{equation*}
    r_t^{\mathrm{net}}(u)
    \le r_t^{\mathrm{loc}}(u)+G\TV_\beta(x_t).
  \end{equation*}
  Next, define the regularized loss by
  \begin{equation*}
    F_t^{\mathrm{reg}}(x)\coloneqq\frac1N\sum_{n=1}^N\ell_{n,t}(x_n)+G\TV_\beta(x).
  \end{equation*}
  Since $\TV_\beta(u\one)=0$, the definition of the regularized loss gives
  \begin{equation*}
    r_t^{\mathrm{loc}}(u)+G\TV_\beta(x_t)
    = F_t^{\mathrm{reg}}(x_t)-F_t^{\mathrm{reg}}(u\one).
  \end{equation*}
  Observe that $\widetilde g_t/N\in\partial F_t^{\mathrm{reg}}(x_t)$, so convexity gives
  \begin{equation*}
    F_t^{\mathrm{reg}}(x_t)-F_t^{\mathrm{reg}}(u\one)
    \le\frac1N\widetilde g_t^\top(x_t-u\one).
  \end{equation*}
  Finally, substituting $a_{n,t}=\lambda_n x_{n,t}$ and $z_{n,t}=\widetilde g_{n,t}/\lambda_n$ yields the one-round reduction
  \begin{equation}
    r_t^{\mathrm{net}}(u)
    \le\frac1N z_t^\top(a_t-u\bm\lambda).
    \label{eq:one-round-reduction}
  \end{equation}
  To conclude, we sum \zcref{eq:one-round-reduction} over time and apply each local regret
  bound with comparator $v=\lambda_n u$.
\end{IEEEproof}

\zcref{thm:main} shows that \zcref{alg:local} is a decentralized algorithm with sublinear comparator-adaptive network regret.
An immediate consequence of \zcref{thm:main} is that if every local regret is sublinear in $T$ for a fixed $v$, then the network regret is sublinear in $T$ for a fixed $u$.

\subsection{Universal weights}\label{sec:universal-weights}

The choice $\beta_e=1/2$ does not require graph preprocessing, that is, each node uses only $G$, $N$, and its degree $d_n$ to compute its scale.
Next, we show that it is a valid choice of weights.
\begin{corollary}[Universal weights]\label{cor:universal-half-bound}\label{lem:universal-half-tv}
  In every connected undirected graph, the weights $\beta_e=1/2$ are admissible and
  \begin{equation*}
    D_N(x)\le\frac12\sum_{\{i,j\}\in E}\abs{x_i-x_j}
    \qquad\text{for every }x\in\R^N.
  \end{equation*}
  The coefficient $1/2$ is the smallest possible uniform bound over all connected graphs.
  With these weights and under \zcref{ass:gradient-bound}, \zcref{alg:local} with base learners satisfying \zcref{def:base-learner} uses the scales
  \begin{equation}
    \lambda_n=G\left(1+\frac N2d_n\right),
    \label{eq:half-tv-scale}
  \end{equation}
  and satisfies
  \begin{equation}
    R_T^{\mathrm{net}}(u)
    \le
    \frac1N\sum_{n=1}^N
    \Reg_{n,T}\!\left(G\left(1+\frac N2d_n\right)u\right).
    \label{eq:half-tv-bound}
  \end{equation}
\end{corollary}
\begin{IEEEproof}
  For every nonempty proper subset $U\subset V$, connectedness gives $\abs{\delta(U)}\ge1$, so
  \begin{equation*}
    \beta(\delta(U))=\frac{\abs{\delta(U)}}2
    \ge\frac12
    \ge\frac{2\abs U(N-\abs U)}{N^2},
  \end{equation*}
  where the last inequality follows from $\abs U(N-\abs U)\le N^2/4$.
  It follows that the weights are admissible, and \zcref{lem:cut-characterization} gives the disagreement bound.

  For sharpness, consider two nodes joined by one edge, and take $x_1\ne x_2$.
  Both sides of the disagreement bound are equal to $\abs{x_1-x_2}/2$, so $1/2$ is sharp uniformly over connected graphs.
  Substituting $\beta_e=1/2$ into the local scales and \zcref{thm:main} gives the regret bound.
\end{IEEEproof}

\section{KT as a concrete instance}\label{sec:kt-analysis}

In this section, we instantiate our black-box reduction with the well-known Krichevsky-Trofimov (KT) algorithm for every local learner~\cite{orabona2016coin,ortega2025deco}.
This algorithm is inspired by coin-betting games, so we will use the original coin-betting terminology.
Note that we present the single-dimensional version of KT here for simplicity, but in the real-data experiments we will use the multi-dimensional extension.
We call this extension radial KT, and we refer the reader to~Ortega and Jafarkhani~\cite[Section~V]{ortega2025deco} for details.

\begin{definition}[KT variants]\label{def:kt-variants}
  Fix an endowment $\eta>0$ and set $S_0=0$.
  For simplicity, we omit the dependence on $\eta$ from the KT notation.
  At round $t\ge1$, predict $a_t$ before receiving feedback $z_t\in[-1,1]$.
  Then, both variants update $S_t=S_{t-1}-z_t$.

  \smallskip\noindent\textbf{Potential-based KT.}
  Predict $a_t=h_t^{\mathrm{KT}}(S_{t-1})$, where
  \begin{align}
    F_t^{\mathrm{KT}}(s)
     & \coloneqq{}
    \eta\frac{2^t
          \Gamma\!\left((t+1+s)/2\right)
          \Gamma\!\left((t+1-s)/2\right)}
        {\pi\Gamma(t+1)},
    \label{eq:kt-potential}                            \\
    h_t^{\mathrm{KT}}(s)
     & \coloneqq{}\frac{s}{t}F_{t-1}^{\mathrm{KT}}(s).
    \label{eq:kt-prediction}
  \end{align}
  The potential is defined for $t\ge0$ and $\abs{s}<t+1$.
  The prediction map is defined for $t\ge1$ and $\abs{s}\le t-1$.

  \smallskip\noindent\textbf{Wealth-based KT.}
  Set the initial wealth $W_0=\eta$.
  Use the current wealth in place of the potential value:
  \begin{equation}
    \begin{aligned}
      a_t & =\frac{W_{t-1}}tS_{t-1}, \\
      W_t & =W_{t-1}-z_t a_t.
    \end{aligned}
    \label{eq:wealth-based-kt}
  \end{equation}
\end{definition}

For the analysis below, we can use either potential-based or wealth-based variant.
To choose the node endowments, define the \emph{average local scale}
\begin{equation*}
  \bar\lambda_\beta
  \coloneqq\frac1N\one^\top\bm\lambda
  =G\left(1+2\sum_{e\in E}\beta_e\right).
\end{equation*}
Fix an average endowment $\eps>0$, and assign the following value to Node $n$
\begin{equation*}
  \eps_n\coloneqq{}\eps\frac{\lambda_n}{\bar\lambda_\beta}.
\end{equation*}
Note that here we have assumed that $\bar \lambda_\beta$ is known to the nodes, which is a mild assumption since it can be pre-computed given a fixed graph topology and edge weights.
We acknowledge that this requires some preprocessing, as previously discussed.
Endowments are designed such that $N^{-1}\sum_n\eps_n=\eps$.
Node $n$ runs either variant from \zcref{def:kt-variants} with endowment $\eps_n$.
With rewards $-z_{n,t}$ and endowment $\eps_n$, the reward-regret relation of Orabona and Pál~\cite[Lemma~1]{orabona2016coin} gives, for every $v\in\R$,
\begin{equation}
  \sum_{t=1}^T z_{n,t}(a_{n,t}-v)
  \le
  \eps_n+F_T^*(v).
  \label{eq:nodewise-kt-regret}
\end{equation}
Here $F_T^*(v)$ is the conjugate of $F_T^{\mathrm{KT}}$, and the wealth-based KT variant satisfies the same bound since $W_T\ge F_T^{\mathrm{KT}}(S_T)$.

\begin{corollary}[Locally scaled KT network regret]\label{cor:kt-network-regret}
  Under \zcref{ass:gradient-bound}, for every $T\ge1$, the proportional-endowment KT instance satisfies
  \begin{equation*}
    R_T^{\mathrm{net}}(u)
    \le
    \bar\lambda_\beta\abs{u}
    \sqrt{T\log\!\left(
      1+\frac{24T^2\bar\lambda_\beta^2u^2}{\eps^2}
      \right)}
    +\eps\left(1-\frac1{e\sqrt{\pi T}}\right).
  \end{equation*}
  The algorithm uses neither $T$ nor $u$.
\end{corollary}
\begin{IEEEproof}
  By \zcref{eq:nodewise-kt-regret}, KT has the local bound $\Reg_{n,T}(v)=\eps_n+F_T^*(v)$.
  For $\eta>0$ and $v\in\R$, Orabona and Pál~\cite[Corollary~5]{orabona2016coin} gives:
  \begin{align*}
    \eta+F_T^*(v)
    \le\abs v\sqrt{T\log\!\left(1+\frac{24T^2v^2}{\eta^2}\right)}+\eta\left(1-\frac1{e\sqrt{\pi T}}\right).
  \end{align*}
  We apply \zcref{thm:main} with $v=\lambda_n u$ and $\eta=\eps_n$.
  Since $\lambda_n/\eps_n=\bar\lambda_\beta/\eps$, the logarithm is the same across nodes, and we can average the local bounds over $n$ to conclude the proof.
\end{IEEEproof}

\section{Why local learning and averaging can fail}\label{sec:counterexamples}

The preceding sections turn local regret bounds into network regret bounds by adding feedback that accounts for prediction disagreement.
To see why this extra feedback matters, we first let each node run KT independently and show that this can lead to catastrophic network regret.
We then add one gossip step to test whether state averaging prevents the failure and show that this also fails.

Throughout this section, we use potential-based KT on a two-node graph, with a common endowment $\eps>0$, subgradients bounded by $G=1$, and comparator $u=0$.
We give the construction of the shared loss here, but leave the proofs to \zcref{app:counterexamples} and the implementation to \zcref{sec:experiments}.

\begin{counterexample}[Two-node kink loss]\label{cex:two-node-kink}
  For both counterexamples, we label nodes $H$, for high, and $L$, for low, with the following subgradients:
  \begin{equation}
    g_{H,t}=-\frac{17}{18},
    \qquad
    g_{L,t}=-\frac1{18}.\label{eq:counterexample-gradients}
  \end{equation}
  For each learner, compute its predictions under these subgradients.
  As shown in \zcref{app:counterexamples}, the predictions satisfy $x_{H,1}=x_{L,1}=0$ and $0<x_{L,t}<x_{H,t}$ for $t\ge2$.
  The high-node loss is
  \begin{equation}
    \ell_{H,t}(x)=-\frac{17}{18}x.\label{eq:counterexample-high-loss}
  \end{equation}
  At the low node, let $\ell_{L,1}(x)=-x/18$ and, for $t\ge2$, place a kink at its prediction:
  \begin{equation}
    \ell_{L,t}(x)=
    \begin{cases}
      -\dfrac1{18}x,
       & x\le x_{L,t}, \\[1.5mm]
      -\dfrac1{18}x_{L,t}+x-x_{L,t},
       & x\ge x_{L,t}.
    \end{cases}\label{eq:kink-loss}
  \end{equation}
  By construction, the losses are convex with unit-bounded subgradients, and the prescribed subgradients are valid at the corresponding predictions.
  A nice feature of this construction is that the losses can be fixed before learning starts, since we can compute all predictions and losses in advance.
  Observe that both losses vanish at $u=0$, and the one-round network regret is
  \begin{equation}
    r_t^{\mathrm{net}}(0)
    =\frac14\sum_{n\in\{H,L\}}\sum_{m\in\{H,L\}}\ell_{m,t}(x_{n,t})
    =\frac{x_{H,t}-37x_{L,t}}{72}.
    \label{eq:kink-one-round-regret}
  \end{equation}
\end{counterexample}

\subsection{Independent KT}

\begin{proposition}[Independent local KT can have exponential network regret]\label{prop:independent-failure}
  Consider two independent scalar potential-based KT learners with endowment $\eps>0$, initial states at zero, predictions $x_{n,t}=h_t^{\mathrm{KT}}(S_{n,t-1})$, and uncorrected updates
  $S_{n,t}=S_{n,t-1}-g_{n,t}$.
  Using the losses and subgradients of \zcref{cex:two-node-kink} with comparator $u=0$, we have
  \begin{equation*}
    \lim_{T\to\infty}\frac{R_T^{\mathrm{net}}(0)}{T}=+\infty.
  \end{equation*}
  Indeed, the cumulative network regret grows exponentially up to polynomial factors.
\end{proposition}

The nodes receive constant negative subgradients of different sizes, so their predictions grow at different exponential rates.
The loss of the lower node has a kink at its prediction, where a negative subgradient is valid.
Beyond the kink, the loss rises with slope one.
Therefore, the higher prediction incurs a large loss at the other node, which makes the cross-evaluation, i.e., the network regret, grow exponentially.
Note that each learner still has its local regret guarantee, but the network regret grows exponentially.

\subsection{KT with one gossip step}

Since the first example does not use communication, we next examine whether averaging neighboring states, i.e., gossiping, prevents the failure.
We use DECO-II, a gossip algorithm that averages the updated states of local learners~\cite{ortega2025deco}.
With potential-based KT and one gossip step per round, its updates are
\begin{equation}
  x_{n,t}=h_t^{\mathrm{KT}}(S_{n,t-1}),
  \qquad
  S_t=M(S_{t-1}-g_t),\label{eq:gossip-baseline-recursion}
\end{equation}
where
\begin{equation}
  M=
  \begin{pmatrix}
    19/20 & 1/20  \\
    1/20  & 19/20
  \end{pmatrix}.\label{eq:counterexample-matrix}
\end{equation}
The disagreement eigenvalue is $9/10$, so the spectral gap is $1/10$.
It follows that each averaging step removes only one tenth of a state difference.
Notice that if we had chosen $M$ with all entries equal to $1/2$, this would remove all disagreement in one step.
Therefore, the choice of weights in \zcref{eq:counterexample-matrix} is needed to make the counterexample work.

\begin{theorem}[DECO-II with one gossip step can have exponential network regret]\label{thm:one-gossip-step-counterexample}
  Fix any endowment $\eps>0$ and run \zcref{eq:gossip-baseline-recursion,eq:counterexample-matrix} from $S_0=0$.
  Using the losses and subgradients of \zcref{cex:two-node-kink} with comparator $u=0$, we have
  \begin{equation*}
    \lim_{T\to\infty}\frac{R_T^{\mathrm{net}}(0)}T=+\infty.
  \end{equation*}
  More precisely, there are constants $c>0$ and $T_0<\infty$, which depend only on $\eps$, such that
  \begin{equation}
    R_T^{\mathrm{net}}(0)
    \ge
    c\eps\frac{\exp\!\left(I(1/2)T\right)}{\sqrt T}
    \qquad(T\ge T_0),\label{eq:gossip-exponential-lower-bound}
  \end{equation}
  where
  \begin{equation*}
    I(s)\coloneqq{}
    \frac{1+s}{2}\log(1+s)
    +
    \frac{1-s}{2}\log(1-s)
  \end{equation*}
  and $I(1/2)>0$.
\end{theorem}

Like before, the nodes receive constant negative subgradients of different sizes.
Gossiping keeps the difference between their states below a constant, while the mean state grows.
As the states grow, the KT map turns the bounded state gap into an exponentially growing prediction gap.
The lower node's loss then penalizes the higher prediction, as in the independent example.
This results in an exponential network regret even though the internal states remain close.
Note that the loss sequence can be fixed before learning starts.

For each fixed number of gossip steps, we can choose an averaging matrix that gives the same failure by simply reducing the problem to the previous case.
The detailed construction and matrices are in \zcref{app:counterexamples}.

\section{Choosing topology-adapted edge weights}\label{sec:graph-factors}

The counterexamples explain why we need our reduction to establish a network regret bound.
We now return to the edge weights and examine how we can design them to minimize their impact on the regret bound.
Setting every edge weight to $1/2$ as in \zcref{cor:universal-half-bound} is simple and does not require graph preprocessing, but it gives a bad average scale,
\begin{equation*}
  \frac1N\sum_n\lambda_n=G(1+\abs E).
\end{equation*}
We can improve this scale by choosing edge-specific weights that are admissible, but we pay the price of needing some additional processing before learning starts.
Next, we discuss several natural constructions and their properties.

\subsection{Path-induced weights}

Our regret bound requires the weighted graph total variation to control the disagreement between every pair of nodes.
However, graph total variation only measures differences between neighboring nodes.
A natural way to connect these two quantities is to choose a path between each pair of nodes and use the triangle inequality along the edges of that path to control the difference between their predictions.
We can flesh out this idea as follows: for each pair of distinct nodes $a,b$, choose a path $P_{ab}$ connecting them and use the same path in reverse for $P_{ba}$.
Let $P_{aa}$ be empty and denote the collection of chosen paths by $\cR$.
The triangle inequality gives
\begin{equation}
  \abs{x_a-x_b}
  \le \sum_{\{i,j\}\in P_{ab}}\abs{x_i-x_j}.
\end{equation}
Next, averaging over all ordered pairs and exchanging the order of summation gives
\begin{align*}
  D_N(x)
   & \le \frac1{N^2}\sum_{a,b}
  \sum_{e=\{i,j\}\in P_{ab}}\abs{x_i-x_j}        \\
   & =\sum_{e=\{i,j\}\in E}b_e^\cR\abs{x_i-x_j},
\end{align*}
where
\begin{equation}
  b_e^\cR
  \coloneqq
  \frac1{N^2}\sum_{a,b}\mathbf1\{e\in P_{ab}\}.
  \label{eq:edge-load}
\end{equation}
The weight $b_e^\cR$ is the fraction of ordered node pairs whose chosen paths use edge $e$.
Observe that edges used by more paths receive larger weights.
Also, by construction, the RHS is our weighted graph total variation, so the weights are admissible by \zcref{lem:cut-characterization}.

The total weight also has a simple interpretation:
\begin{equation*}
  \sum_{e\in E}b_e^\cR
  =\frac1{N^2}\sum_{a,b}|P_{ab}|,
\end{equation*}
which is the average length of the chosen paths.
It follows that to minimize the total weight, we must choose the shortest paths between each pair of nodes.

\subsection{Two edge-weight objectives}

Observe that the choice of weights determines the local scales $\lambda_n=G(1+\kappa_n)$.
Therefore, we seek weights that keep these scales small while satisfying the cut constraints.
Next, we consider two objectives: for the first, we minimize the maximum local scale, which limits the scaling required at any node; for the second, we minimize the average local scale, which directly improves the network regret bound with some learners, like the KT learner we introduced in \zcref{sec:kt-analysis}.
Formally, let $\cC(\cG)$ denote the set of all admissible weights:
\begin{equation*}
  \cC(\cG)
  \coloneqq
  \left\{\beta\ge0:
  \begin{array}{l}
    \beta(\delta(U))\ge
    \dfrac{2\abs U(N-\abs U)}{N^2} \\
    \text{for every }\varnothing\ne U\subsetneq V
  \end{array}
  \right\}.
\end{equation*}
The corresponding optimization problems are
\begin{align*}
  \min_{\beta\in\cC(\cG)}
  \max_{n\in V}\kappa_n, \\
  \min_{\beta\in\cC(\cG)}
  2\norm{\beta}_1.
\end{align*}
The first minimizes the \emph{maximum incident load} $\norm{\bm\kappa}_\infty=\max_{n\in V}\kappa_n$.
This also minimizes the \emph{maximum local scale}
\begin{equation*}
  \lambda_{\max}
  \coloneqq\max_{n\in V}\lambda_n
  =G(1+\norm{\bm\kappa}_\infty).
\end{equation*}

The second minimizes the total edge weight, since $\norm{\beta}_1=\sum_{e\in E}\beta_e$ for nonnegative weights.
We call admissible weights with the smallest total edge weight \emph{mass-optimal}.
Because
\begin{equation*}
  \bar\lambda_\beta
  =G\left(1+2\norm{\beta}_1\right).
\end{equation*}
This also minimizes the average local scale that appears in the KT regret bound.

Both problems can be written as linear programs.
However, they have one constraint for each nonempty proper subset of nodes.
Thus, the number of constraints grows exponentially with $N$.
We can solve these programs for the small graphs in our experiments, but for larger graphs, we can use path-based weights, or other weight constructions, as discussed in the next subsection.

\subsection{Trees, complete graphs, and normalization factors}

On a tree, removing an edge $e$ divides the graph into components of sizes $s_e$ and $N-s_e$.
Since $e$ is the only edge that crosses this cut, admissibility requires
\begin{equation}
  \beta_e\ge\frac{2s_e(N-s_e)}{N^2}.
  \label{eq:tree-edge-load}
\end{equation}
There is exactly one path between any two nodes.
This path uses $e$ precisely when the nodes belong to different components.
There are $2s_e(N-s_e)$ such ordered pairs, so the path-induced weights from \zcref{eq:edge-load} achieve equality in \zcref{eq:tree-edge-load} on every edge.
Note that these weights are admissible, and no admissible choice can reduce any individual edge weight.
Therefore, these weights minimize both the maximum incident load and the total edge weight.

For these weights, suppose that removing Node $n$ leaves components of sizes $a_{n,1},\ldots,a_{n,d_n}$.
Since these sizes sum up to $N-1$, the local incident load is
\begin{align}
  \kappa_n
   & =\frac2N\sum_{r=1}^{d_n}a_{n,r}(N-a_{n,r}) \notag \\
   & =2(N-1)-\frac2N\sum_{r=1}^{d_n}a_{n,r}^2.
  \label{eq:tree-local-load}
\end{align}

On a complete graph, the optimal weights $\beta_e=2/N^2$ give equality in every cut constraint.
In particular, each node has the smallest incident load allowed by its single-node cut.
These weights therefore minimize both objectives and give
\begin{equation*}
  \lambda_n=\lambda_{\max}=\bar\lambda_\beta=G(3-2/N).
\end{equation*}

\begin{table}[t]
  \centering
  \caption{Maximum and average factors for edge-weight constructions.
    Here $d_{\max}\coloneqq\max_{n\in V}d_n$ is the maximum node degree.}\label{tab:graph-factors}
  \footnotesize
  \resizebox{\columnwidth}{!}{%
    \begin{tabular}{@{}llll@{}}
      \toprule
      Construction & \shortstack{Positive-weight\\edges} & $1+\norm{\bm\kappa}_\infty$ & $1+2\sum_e\beta_e$ \\
      \midrule
      Complete, optimal
                   & $\binom N2$                         & $3-2/N$                     & $3-2/N$            \\
      Path, $N$ even
                   & $N-1$                               & $N+1-2/N$                   & $1+2(N^2-1)/(3N)$  \\
      Path, $N$ odd
                   & $N-1$                               & $N+1-1/N$                   & $1+2(N^2-1)/(3N)$  \\
      Star
                   & $N-1$                               & $1+2{(N-1)}^2/N$            & $1+4{(N-1)}^2/N^2$ \\
      Universal half-TV
                   & $\abs E$                            & $1+Nd_{\max}/2$             & $1+\abs E$         \\
      \bottomrule
    \end{tabular}%
  }
\end{table}
\zcref{tab:graph-factors} summarizes the maximum and average factors for several standard graph constructions, including path and star graphs.
Multiplying the last two columns by $G$ gives the maximum local scale $\lambda_{\max}$ and the average local scale $\bar\lambda_\beta$.
Observe that universal half-TV weights $\beta_e=1/2$ work on every connected graph and do not require graph preprocessing, but they can be far from optimal.
For example, on a complete graph, the optimal weights reduce both factors from $1+N(N-1)/2$ to $3-2/N$.

\section{Extension to Hilbert spaces}\label{sec:extensions}

We now extend the scalar reduction to predictions in a real Hilbert space $\cH$.
Throughout this section, the edge weights and local scales remain the same.
To simplify the exposition, we fix a round $t$ and omit $t$ from the notation.
As in the scalar case, assume that each local loss is convex and has a subgradient at every point.
Now, the subgradient assumption becomes
\begin{equation*}
  \sup_{x\in\cH}\sup_{g\in\partial\ell_{n,t}(x)}\norm g\le G.
\end{equation*}
For corrections, we simply replace the scalar sign with the unit vector along each prediction difference.
Then, the correction at Node $n$ becomes
\begin{equation*}
  c_n
  =GN\sum_{j:\{n,j\}\in E}\beta_{\{n,j\}}
  \frac{x_n-x_j}{\norm{x_n-x_j}},
\end{equation*}
where the contribution is zero when $x_n=x_j$.
For every comparator $u\in\cH$, these corrections satisfy
\begin{align*}
  \sum_n c_n
   & =0,                                                 \\
  \frac1N\sum_n\inner{c_n}{x_n-u}
   & =G\sum_{\{i,j\}\in E}\beta_{\{i,j\}}\norm{x_i-x_j}, \\
  \norm{c_n}
   & \le G\kappa_n.
\end{align*}

Next, we show that the disagreement bound also holds with norms.
If all predictions are zero, the bound is immediate.
Otherwise, let $\mathcal V=\operatorname{span}\{x_1,\ldots,x_N\}$.
Choose a standard Gaussian vector $Z\sim\mathcal N(0,I_{\mathcal V})$.
Given admissible weights $\beta$, we apply \zcref{lem:cut-characterization} to the scalar projections $\inner{Z}{x_n}$:
\begin{equation*}
  \frac1{N^2}\sum_{n,m}\abs{\inner{Z}{x_n-x_m}}
  \le
  \sum_{\{i,j\}\in E}\beta_{\{i,j\}}\abs{\inner{Z}{x_i-x_j}}.
\end{equation*}

For every $v\in\mathcal V$, the projection $\inner{Z}{v}$ is Gaussian with mean zero and variance $\norm v^2$.
Therefore,
\begin{equation*}
  \mathbb E_Z\abs{\inner{Z}{v}}
  =\sqrt{\frac2\pi}\norm v.
\end{equation*}
We take expectations in the projected inequality and cancel $\sqrt{2/\pi}$ to obtain the disagreement bound in the Hilbert space:
\begin{equation*}
  \frac1{N^2}\sum_{n,m}\norm{x_n-x_m}
  \le
  \sum_{\{i,j\}\in E}\beta_{\{i,j\}}\norm{x_i-x_j}.
\end{equation*}
Therefore, \zcref{thm:main} also holds for base learners with predictions in a Hilbert space.

\section{Experiments}\label{sec:experiments}

We present experiments on the two counterexample constructions presented in~\zcref{sec:counterexamples}, we compare edge weight constructions on synthetic data, and we evaluate radial KT on real data.\footnote{Code: \url{https://github.com/TomasOrtega/comparator-adaptive-network-regret}}
We give supplementary details in \zcref{app:experiments}.

\subsection{Adaptive independent-KT stress test}\label{sec:adaptive-kink-experiment}

We apply the loss construction of \zcref{cex:two-node-kink} to wealth-based KT for $300$ rounds with $\eps=1$ and the comparator $u=0$.
The corrected learner uses the single edge with $\beta_e=1/2$.

Due to space constraints, we do not plot the losses, but independent KT ends with network regret $\KinkIndependentTerminal$, while edge-corrected KT ends with network regret $\KinkCorrectedTerminal$.
So, we conclude that the correction prevents the failure, as predicted in \zcref{prop:independent-failure}.

\subsection{Fixed counterexample with one gossip step}\label{sec:gossip-counterexample-experiment}

\begin{figure}[htbp]
  \centering
  \includegraphics[width=\linewidth]{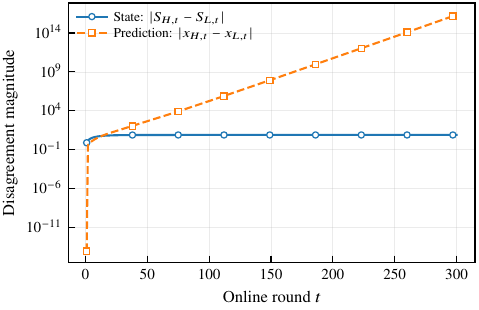}
  \caption{State and prediction disagreement with one gossip step.
    The exact state gap stays below eight, while the potential-based KT prediction gap grows exponentially.}\label{fig:gossip-state-prediction}
\end{figure}

As plotted in \zcref{fig:gossip-state-prediction}, we reproduce \zcref{thm:one-gossip-step-counterexample} for $\GossipCounterexampleHorizon$ rounds with the comparator $u=0$.
Note that all four plotted methods receive the same fixed sequence of losses.
All methods use the potential-based KT map in \zcref{eq:kt-prediction}, and DECO-II/KT applies one gossip step per round.

As predicted by our theory, \zcref{fig:gossip-state-prediction} shows why bounded state disagreement does not control the prediction disagreement.
DECO-II/KT ends with a network regret of $\GossipBaselineTerminal$ and a cumulative average local regret of $\GossipBaselineLocalTerminal$.
Edge-corrected KT ends with a negative network regret on the same losses.
\zcref{tab:gossip-counterexample} provides the detailed final results.
\begin{table}[htbp]
  \centering
  \caption{Terminal results for the fixed counterexample.
    Prediction spread is $\max_n x_{n,T}-\min_n x_{n,T}$.}\label{tab:gossip-counterexample}
  \resizebox{\columnwidth}{!}{%
\begin{tabular}{@{}lrrr@{}}
\toprule
Method & Network regret & Local regret & Prediction spread \\
\midrule
Edge-corrected KT & $-1161.311$ & $-1228.614$ & $25.819$ \\
DECO-II/KT (one gossip step) & $1.391\times10^{15}$ & $-8.334\times10^{16}$ & $2.108\times10^{16}$ \\
Independent KT & $4.664\times10^{70}$ & $-1.586\times10^{72}$ & $1.449\times10^{72}$ \\
Centralized KT oracle & $-9.141\times10^{14}$ & $-9.141\times10^{14}$ & $0$ \\
\bottomrule
\end{tabular}
}
\end{table}

\subsection{Heterogeneous synthetic absolute losses}\label{sec:synthetic-absolute-experiment}

We use $N=8$ nodes and losses $\ell_{n,t}(x)=\abs{x-y_{n,t}}$ for $\ExperimentHorizon$ rounds over $\ExperimentSeeds$ seeds.
We develop a synthetic sequence, described in \zcref{eq:synthetic}, with different node means, a common sinusoidal drift, and Gaussian noise.
All KT methods in this experiment are wealth-based, and corrected methods use proportional endowments with $\eps=1$.

\begin{figure}[htbp]
  \centering
  \includegraphics[width=\linewidth]{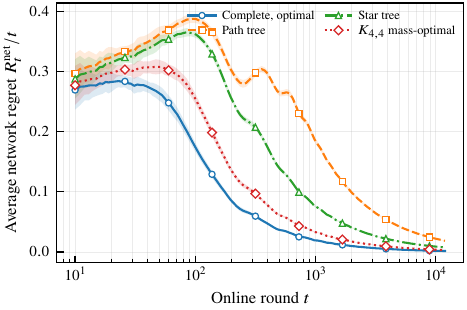}
  \caption{Average network regret on synthetic absolute losses, with nodewise scales.
    Curves show means over $\ExperimentSeeds$ seeds; bands show two standard errors.}\label{fig:absolute-topologies}
\end{figure}

\zcref{fig:absolute-topologies} compares four communication graphs to show the effect of network topology.
The complete graph has the lowest terminal regret and requires the most communication.
Observe that the star has lower regret compared to the path despite its larger maximum node load.

\begin{figure}[htbp]
  \centering
  \includegraphics[width=\linewidth]{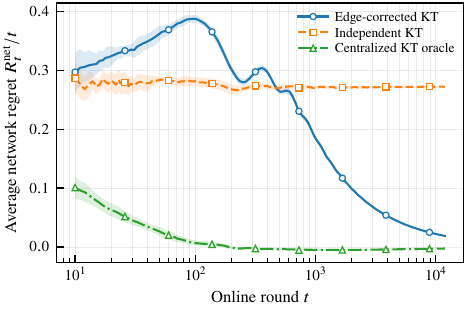}
  \caption{Path-based edge-corrected KT, independent KT, and centralized KT on the synthetic data.
    Bands show two standard errors.}\label{fig:absolute-methods}
\end{figure}

With the same data, we also compare path-based edge-corrected KT with independent KT, and centralized KT in \zcref{fig:absolute-methods}.
It shows that our correction helps on data beyond the adversarial counterexamples, as independent KT performs worse than edge-corrected KT, trending toward the centralized KT baseline.

Finally, we compare different edge-weight constructions on $K_{4,4}$.
We construct path-induced weights from the shortest paths using \zcref{eq:edge-load}.
\zcref{tab:experimental-results} shows that the mass-optimal weights provide the lowest average scale and terminal regret among these choices.

\begin{table}[htbp]
  \centering
  \caption{Edge-corrected KT on synthetic losses over $\ExperimentSeeds$ seeds at $T=\ExperimentHorizon$.
    Cumulative regret gives the mean and sample standard deviation.
    Per-round regret gives the mean.
    Scales use $G=1$.}\label{tab:experimental-results}
  \footnotesize
  \setlength{\tabcolsep}{3pt}
\begin{tabular}{@{}p{0.34\columnwidth}rrrr@{}}
\toprule
Graph / weights & $\bar{\lambda}_{\beta}$ & $\lambda_{\max}$ & $R_T^{\mathrm{net}}$ & $R_T^{\mathrm{net}}/T$ \\
\midrule
Complete, optimal & 2.75 & 2.75 & 20.8 $\pm$ 0.9 & 0.0017 \\
Path tree & 6.25 & 8.75 & 225.3 $\pm$ 5.1 & 0.0188 \\
Star tree & 4.06 & 13.25 & 92.5 $\pm$ 1.4 & 0.0077 \\
$K_{4,4}$ path-induced & 3.50 & 5.75 & 59.2 $\pm$ 1.6 & 0.0049 \\
$K_{4,4}$ mass-optimal & 3.00 & 3.00 & 39.0 $\pm$ 1.1 & 0.0032 \\
$K_{4,4}$ uniform half weights & 17.00 & 17.00 & 3208.0 $\pm$ 24.7 & 0.2673 \\
\bottomrule
\end{tabular}

\end{table}

\subsection{Real-world online experiments}\label{sec:real-world-experiment}

We use four LIBSVM datasets: \texttt{cadata}, \texttt{cpusmall}, \texttt{space\_ga}, and \texttt{abalone}~\cite{chang2011libsvm,libsvmregressiondata,ortega2025deco}.
For each of $\RealWorldSeeds$ seeds, we shuffle the data and assign one example to each of $\RealWorldNodes$ nodes per round.
We discard the final incomplete batch.
We append an intercept and normalize each feature row to have the norm at most one, so $G=1$.
Our network loss is
\begin{equation*}
  L_t^{\mathrm{net}}\coloneqq{}
  \frac1{N^2}\sum_{n=1}^N\sum_{m=1}^N
  \abs{\inner{w_{n,t}}{\phi_{m,t}}-y_{m,t}}.
\end{equation*}

We compare the edge-corrected radial KT on $K_{5,5}$ and the complete graph.
Edge-corrected, independent, and centralized radial KT use the wealth-based variant.
DECO-II/KT uses potential-based radial KT with one gossip step.
The $K_{5,5}$ weights minimize the maximum node load.
All KT methods use endowment $\eps=\RealWorldEndowment$.

We also compare D-OGD and coordinatewise D-AdaGrad~\cite{zinkevich2003online,yan2013distributed,duchi2011adaptive}.
DECO-II/KT, D-OGD, and D-AdaGrad each use one gossip step on the same $K_{5,5}$ graph as edge-corrected KT.\@
For D-OGD, each swept value $\eta_0$ sets the step size $\eta_t=\eta_0/\sqrt t$.

\begin{figure*}[htbp]
  \centering
  \includegraphics[width=0.92\textwidth]{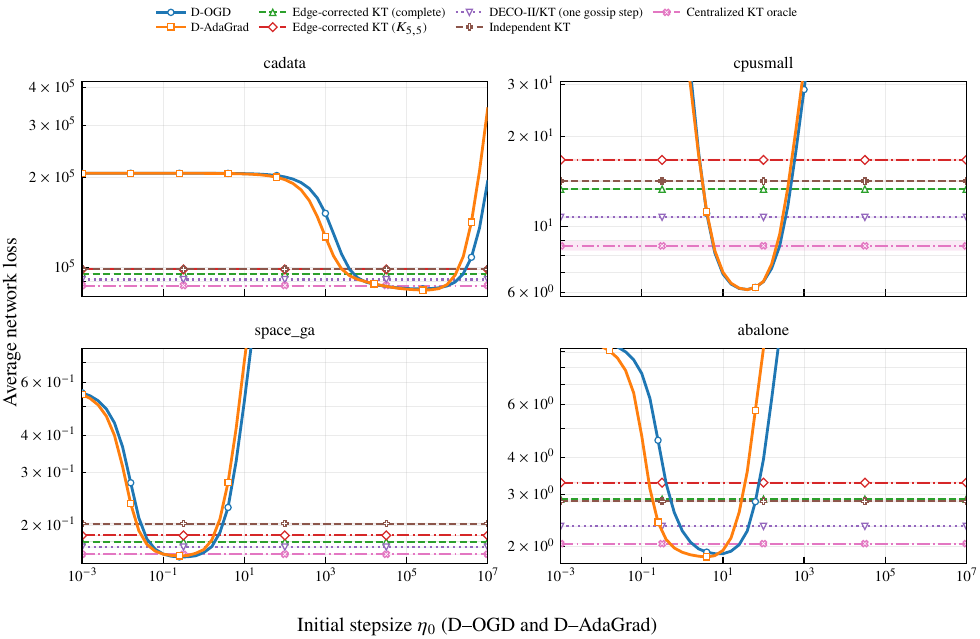}
  \caption{Average network loss, $T^{-1}\sum_{t=1}^T L_t^{\mathrm{net}}$, versus initial step size for D-OGD and D-AdaGrad.
    Fixed KT configurations give horizontal lines.
    The $y$-axis is truncated to show the relevant range.
    Larger values lie outside the displayed range.
    Bands show two standard errors over $\RealWorldSeeds$ permutations.}\label{fig:real-world-loss}
\end{figure*}

\zcref{fig:real-world-loss} shows the performance of D-OGD and D-AdaGrad across initial step sizes from $10^{-3}$ to $10^7$.
At their best swept steps, both gradient methods have a lower average network loss compared to every fixed KT method for each dataset.
Observe that DECO-II/KT has a lower average network loss compared to edge-corrected KT on $K_{5,5}$ and on the complete graph for all four datasets.
This is an indication that it might have empirical advantages despite not guaranteeing the worst-case performance like edge-corrected methods.

\section{Conclusion}\label{sec:conclusion}

We have shown that, given local learners with comparator-adaptive regret guarantees, we can construct a network learner with a comparator-adaptive network regret guarantee.
To ease analysis, we have characterized the edge weights that are admissible for the disagreement correction, and have shown that the choice of edge weights can significantly affect the regret bound.
To guide practical implementations, we have given design choices for the edge weights and shown that a universal choice of edge weights, although not optimal, can be used without any graph preprocessing.
Experiments on synthetic and real-world data support our theoretical results.

\appendices{}
\zcsetup{countertype={section=appendix}}

\section{Counterexample proofs}\label{app:counterexamples}

We use the counterexample construction from \zcref{cex:two-node-kink}, and use \zcref{eq:kink-one-round-regret} to bound their regret.

\subsection{Independent KT without correction}\label{app:independent-failure}

\begin{IEEEproof}[Proof of \zcref{prop:independent-failure}]
  Under \zcref{eq:counterexample-gradients}, the states before round $t$ are
  \begin{equation*}
    S_{H,t-1}=\frac{17}{18}(t-1),
    \qquad
    S_{L,t-1}=\frac1{18}(t-1),
  \end{equation*}
  so $0<x_{L,t}<x_{H,t}$ for $t\ge2$.
  The growth estimate in \zcref{eq:kt-growth}, with the rate function $I$ defined there, gives
  \begin{equation*}
    \frac{x_{H,t}}{x_{L,t}}
    =\exp\!\left(t\left[I(17/18)-I(1/18)\right]+O(\log t)\right)
    \longrightarrow\infty,
  \end{equation*}
  because $I'(s)=\operatorname{atanh}(s)>0$ on $(0,1)$.
  Eventually $x_{H,t}\ge74x_{L,t}$, so \zcref{eq:kink-one-round-regret} gives $r_t^{\mathrm{net}}(0)\ge x_{H,t}/144$.
  The same estimate in \zcref{app:kt-asymptotics} shows that $x_{H,t}$ grows exponentially up to a polynomial factor, as desired.
\end{IEEEproof}

\subsection{DECO-II with one gossip step}\label{app:gossip-failure}

Now, let us write the mean state and half of the state difference as
\begin{equation*}
  m_t\coloneqq{}\frac{S_{H,t}+S_{L,t}}2,
  \qquad
  d_t\coloneqq{}\frac{S_{H,t}-S_{L,t}}2.
\end{equation*}
Use the same decomposition for the subgradients:
\begin{equation*}
  \bar g_t\coloneqq{}\frac{g_{H,t}+g_{L,t}}2,
  \qquad
  \delta_t\coloneqq{}\frac{g_{H,t}-g_{L,t}}2.
\end{equation*}
Direct substitution into \zcref{eq:gossip-baseline-recursion} gives
\begin{equation}
  m_t=m_{t-1}-\bar g_t,
  \qquad
  d_t=\frac9{10}(d_{t-1}-\delta_t).\label{eq:mean-disagreement-recursion}
\end{equation}
Observe that the mean state follows the centralized update, while the averaging step multiplies the state difference by $9/10$.

\begin{IEEEproof}[Proof of \zcref{thm:one-gossip-step-counterexample}]
  Under \zcref{eq:counterexample-gradients}, $\bar g_t=-1/2$ and $\delta_t=-4/9$.
  Starting from $m_0=d_0=0$, \zcref{eq:mean-disagreement-recursion} has the exact solution
  \begin{equation}
    m_t=\frac t2,
    \qquad
    d_t=4\left(1-{\left(\frac9{10}\right)}^t\right).\label{eq:gossip-exact-state}
  \end{equation}
  Therefore, the following two states generate the predictions at round $t$:
  \begin{align}
    S_{H,t-1}
     & =\frac{t-1}{2}
    +4\left(1-{\left(\frac9{10}\right)}^{t-1}\right),\label{eq:gossip-high-state} \\
    S_{L,t-1}
     & =\frac{t-1}{2}
    -4\left(1-{\left(\frac9{10}\right)}^{t-1}\right).\label{eq:gossip-low-state}
  \end{align}
  We can check that the state disagreement is smaller than eight.
  Now, let
  \begin{equation*}
    x_{H,t}\coloneqq{}h_t^{\mathrm{KT}}(S_{H,t-1}),
    \qquad
    x_{L,t}\coloneqq{}h_t^{\mathrm{KT}}(S_{L,t-1}).
  \end{equation*}
  For every $t\ge2$, we have $0<x_{L,t}<x_{H,t}$.

  The KT estimates in \zcref{app:kt-asymptotics} give
  \begin{align*}
    \frac{x_{H,t}}{x_{L,t}} & \longrightarrow81,                                   \\
    x_{L,t}                 & \ge C\eps\frac{\exp\!\left(tI(1/2)\right)}{\sqrt t}.
  \end{align*}
  Here, $C>0$ and the second inequality holds for all sufficiently large $t$.
  Therefore, substituting into \zcref{eq:kink-one-round-regret} gives
  \begin{equation*}
    \frac{r_t^{\mathrm{net}}(0)}{x_{L,t}}
    \longrightarrow\frac{81-37}{72}=\frac{11}{18}>\frac14,
  \end{equation*}
  and $r_t^{\mathrm{net}}(0)\ge x_{L,t}/4$ for all sufficiently large $t$.

  The finitely many rounds before these inequalities hold contribute a fixed finite amount, and all later rounds have positive regret.
  Since $I(1/2)>0$, the last round's regret eventually exceeds twice the magnitude of the initial contribution.
  This gives the stated cumulative lower bound with $c=C/8$.
\end{IEEEproof}

\paragraph{A fixed number of gossip steps}
Now, consider the case where DECO-II/KT uses a fixed number $s\ge1$ of gossip steps per round.
We can adapt \zcref{cex:two-node-kink} to also rule out this case.
Let $\eta_s={(9/10)}^{1/s}$ and choose
\begin{equation*}
  M_s\coloneqq\frac12
  \begin{pmatrix}
    1+\eta_s & 1-\eta_s \\
    1-\eta_s & 1+\eta_s
  \end{pmatrix}.
\end{equation*}
Its eigenvalues are $1$ and $\eta_s$, so $M_s^s$ equals the counterexample matrix in \zcref{eq:counterexample-matrix}.
Therefore, DECO-II/KT with $s$ gossip steps per round has the same updates and exponential network regret as in \zcref{thm:one-gossip-step-counterexample}.
This concludes he proof.

\section{KT asymptotics for the counterexamples}\label{app:kt-asymptotics}

The following standard consequences of Stirling's formula contain the calculations for \zcref{prop:independent-failure,thm:one-gossip-step-counterexample}.
Uniformly for $a$ in a compact set,
\begin{equation}
  \begin{aligned}
    \log\Gamma(y+a)
      &=\left(y+a-\frac12\right)\log y-y \\
      &\quad+\frac12\log(2\pi)+O(y^{-1})
        \qquad(y\to\infty).
  \end{aligned}
  \label{eq:uniform-stirling}
\end{equation}
Therefore, whenever $a_y,b_y$ converge to finite limits,
\begin{equation*}
  \frac{\Gamma(y+a_y)}{\Gamma(y+b_y)}
  =y^{a_y-b_y}(1+o(1)).
\end{equation*}

Fix an endowment $\eps>0$, a slope $s\in(0,1)$, and a bounded sequence $b_t$.
We evaluate the KT map at the state $s_t=st+b_t$ in \zcref{eq:kt-potential,eq:kt-prediction}.
Another application of \zcref{eq:uniform-stirling} yields
\begin{equation}
  \log h_t^{\mathrm{KT}}(st+b_t)
  =tI(s)-\frac12\log t+O(1),
  \label{eq:kt-growth}
\end{equation}
where
\begin{equation*}
  I(s)=\frac{1+s}{2}\log(1+s)
  +\frac{1-s}{2}\log(1-s).
\end{equation*}
The remainder is uniform when $b_t$ stays in a fixed compact interval.
Since $I'(s)=\operatorname{atanh}(s)>0$, this proves the exponential separation used in \zcref{prop:independent-failure}.
For the low state in \zcref{eq:gossip-low-state}, $s=1/2$, and therefore
\begin{equation}
  \log x_{L,t}=tI(1/2)-\frac12\log t+O(1).
  \label{eq:gossip-low-growth}
\end{equation}

For fixed $b_1,b_2$, the same calculation gives
\begin{equation*}
  \frac{h_t^{\mathrm{KT}}(st+b_1+o(1))}{h_t^{\mathrm{KT}}(st+b_2+o(1))}
  \longrightarrow
  \exp\!\left((b_1-b_2)\operatorname{atanh}s\right).
\end{equation*}
For \zcref{eq:gossip-high-state,eq:gossip-low-state}, $s=1/2$ and $b_1-b_2=8$, so
\begin{equation}
  \frac{x_{H,t}}{x_{L,t}}\longrightarrow
  \exp\!\left(8\operatorname{atanh}(1/2)\right)=81.
  \label{eq:gossip-ratio-limit}
\end{equation}

\section{Experimental protocol details}\label{app:experiments}

\paragraph{KT updates}
Corrected, independent, and centralized KT use the wealth-based variant in \zcref{def:kt-variants}, except in the fixed counterexample.
For corrected KT, $a_t$ is the unscaled action, $z_t$ is normalized augmented feedback, and $\eta=\eps_n$.
Centralized KT uses the mean local subgradient.
All four methods use the potential-based KT in the fixed counterexample.
On real data, DECO-II/KT uses the potential-based radial KT.\@

\paragraph{Synthetic heterogeneous stream}
The absolute-loss suite uses seeds $0,\ldots,4$.
For $n=1,\ldots,8$ and $t=1,\ldots,T$, we generate
\begin{equation}
  \begin{aligned}
    y_{n,t} & =\mu_n+0.25\sin\!\left(\frac{2\pi(t-1)}{250}\right)+0.30\xi_{n,t}, \\
    \mu_n   & =-0.5+\frac{3(n-1)}7.
  \end{aligned}
  \label{eq:synthetic}
\end{equation}
The variables $\xi_{n,t}$ are independent standard normal draws.
For each seed, we use the median of all $8T$ data points as the comparator.

\paragraph{Real data}
Preprocessing adds an intercept and divides each row by the larger of its Euclidean norm and one.
We fit one least-absolute-deviation comparator to each dataset by median regression~\cite{koenker2005quantile}, which we only use to compute the regret.
\zcref{tab:real-datasets} gives the size, dimension, horizon, and reference loss for each dataset.

\begin{table}[H]
  \centering
  \caption{The dimension includes the intercept.
    Reference losses are averages in the original label units.}\label{tab:real-datasets}
  \footnotesize
  \resizebox{\columnwidth}{!}{%
\begin{tabular}{@{}lrrrrr@{}}
\toprule
Dataset & Examples & Dimension & Rounds & Median loss & Comparator loss \\
\midrule
cadata & 20,640 & 9 & 2,064 & 88354.131 & 71283.889 \\
cpusmall & 8,192 & 13 & 819 & 9.661 & 3.737 \\
space\_ga & 3,107 & 7 & 310 & 0.153 & 0.096 \\
abalone & 4,177 & 9 & 417 & 2.359 & 1.550 \\
\bottomrule
\end{tabular}
}
\end{table}

DECO-II/KT, D-OGD, and D-AdaGrad use the symmetric Metropolis matrix of $K_{5,5}$.

D-OGD uses step size $\eta_0/\sqrt t$.
For D-AdaGrad, the step size at Node $n$ in coordinate $j$ is
\begin{equation*}
  \frac{\eta_0}{\sqrt{10^{-8}+\sum_{s=1}^t g_{n,s,j}^2}}.
\end{equation*}
Both methods update locally before averaging.

\section*{Acknowledgment}
OpenAI Codex helped revise the text, write the code for the experiments, and find the counterexamples.
The authors take full responsibility for the work.

\bibliography{references}
\bibliographystyle{IEEEtran}
\end{document}